\documentclass[numbers]{article}
\pdfoutput=1

\usepackage[final]{neurips_2021}

\usepackage[utf8]{inputenc} 
\usepackage[T1]{fontenc}    
\usepackage{hyperref}       
\usepackage{url}            
\usepackage{booktabs}       
\usepackage{amsfonts}       
\usepackage{nicefrac}       
\usepackage{microtype}      
\usepackage{xcolor}         
\usepackage{graphicx}              
\usepackage{multirow}
\usepackage{comment}
\usepackage{subcaption}
\usepackage{makecell}
\usepackage{caption}
\usepackage{comment}
\graphicspath{{figures/}} 

\title{Privacy Aware Person Detection in Surveillance Data}

\author{%
  Sander De Coninck \\
  Ghent University - imec\\
  \texttt{Sander.DeConinck@UGent.be} \\
  \And
   Sam Leroux \\
  Ghent University - imec \\
  \texttt{Sam.Leroux@UGent.be} \\
  \And
   Pieter Simoens \\
   Ghent University - imec \\
   \texttt{Pieter.Simoens@UGent.be} \\
}

\begin{document}

\maketitle

\begin{abstract}
    Crowd management relies on inspection of surveillance video either by operators or by object detection models. These models are large, making it difficult to deploy them on resource constrained edge hardware. Instead, the computations are often offloaded to a (third party) cloud platform. While crowd management may be a legitimate application, transferring video from the camera to remote infrastructure may open the door for extracting additional information that are infringements of privacy, like person tracking or face recognition. In this paper, we use adversarial training to obtain a lightweight obfuscator that transforms video frames to only retain the necessary information for person detection. Importantly, the obfuscated data can be processed by publicly available object detectors without retraining and without significant loss of accuracy.
\end{abstract}
\section{Introduction}
In the past years, we have seen a staggering amount of surveillance cameras deployed in our streets, shops and cities~\cite{WSJPrivacy}. While these cameras are installed for legitimate reasons, there is an intrinsic risk that the video data will be used for unintended purposes~\cite{cavailaro2007privacy}. Reducing this risk of function creep is especially important when the video analysis is offloaded from the camera to a third party platform using a Software as a Service (SaaS) model. Using SaaS is often preferred as object detection can be too computationally expensive for edge devices and third party software gives better chances at state of the art results. In this paper, we obfuscate surveillance video frames to hide identity revealing information while still allowing for accurate person detection. Crucially, our method of data obfuscation does not require adaptations or retraining of the downstream detector models and generalizes across cameras and object detectors.

We use adversarial training to obtain an obfuscator model to transform video frames in such a way that object detection can still be performed with high accuracy but no other tasks. Furthermore, we propose to do this obfuscation using a computationally efficient model that is suitable for execution on the camera. We then transmit the obfuscated data to a cloud backend to perform the computationally expensive object detection. 

We focus on inference phase privacy, thereby assuming access to a pretrained object detection model. A powerful technique for this is Homomorphic Encryption (HE) which can perform calculations on encrypted data without ever decrypting it ~\cite{pmlr-v48-gilad-bachrach16, juvekar2018gazelle}. A disadvantage of HE is its large computational cost~\cite{epic}.

An alternative technique is called Secure Multi-party Computation (SMC). It ensures privacy by splitting computations over several computing parties~\cite{miniONN}. SMC is mainly fit for simple functions, usage for complex functions such as neural networks is possible but similarly to HE often results in heavy computation costs~\cite{epic}.

A promising category in inference phase privacy are data transformation techniques. Their main goal is to limit the transmitted data to the bare minimum required for the task. Research includes Leroux et al.'s~\cite{leroux2018privacy} adversarial obfuscator which forms the basis for this work and Mireshghallah's Shredder~\cite{mireshghallah2020shredder}. Shredder splits a network in two parts in which one part runs on the cloud and the other on a users device. Before sending data from the users device to the cloud, noise is added to ensure privacy. A disadvantage of Shredder is that it requires modification of the downstream model. 

Other work focuses specifically on privacy aware people detection through face anonymization~\cite{face_anon, liuvisual}. This however, requires an often computationally expensive face detection method to be executed on the edge device and could possibly still allow a persons identity to be retrieved due to other attributes (clothing, sex, haircut) remaining visible~\cite{re-id}.

In this paper, we build upon our previous work ~\cite{leroux2018privacy}. Our current contribution scales the technique up to real-world surveillance data and provides experimental benchmarking with other obfuscation methods. Furthermore, we assess its generalization performance across multiple cameras and across object detector models.

\section{Privacy preserving obfuscation}
\label{sec:arch}
\begin{figure}[h]
\begin{center}
  \includegraphics[width=\textwidth]{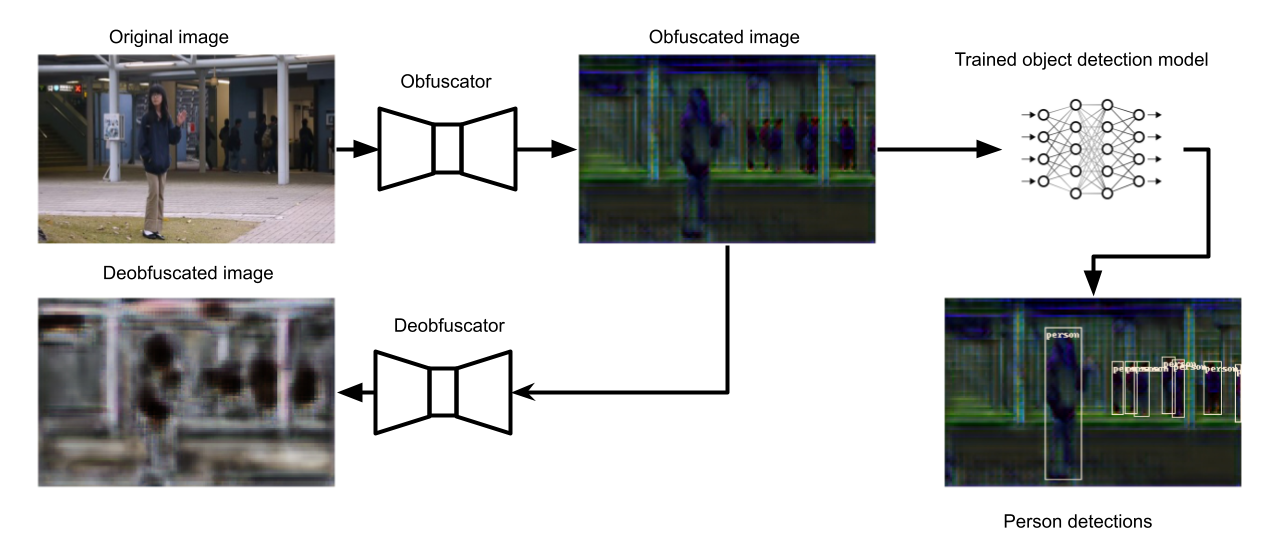}
  \caption{We use an adversarial approach to train an obfuscator autoencoder model.}
  \label{fig:advobf}
\end{center}
\end{figure}

Our architecture is shown in Figure \ref{fig:advobf} and  consists of an obfuscator and deobfuscator model that are trained in opposition to each other, similar to Generative Adversarial Networks (GANs)~\cite{goodfellow2014generative}. The obfuscator learns to transform frames in such a way that the deobfuscator cannot reconstruct them while ensuring a pretrained object detection model can still make accurate predictions. The deobfuscator's only goal is to reconstruct obfuscated frames to their original form. At test time, the deobfuscator is discarded. Details on the training procedure can be found in the appendix.

The obfuscator and deobfuscator are both autoencoders based on the MobileNet \cite{howard2017mobilenets} architecture. This architecture is designed for lightweight and fast models to be executed on resource constrained devices.

First and foremost, the goal of our technique is to obfuscate video frames in such a way that their contents are no longer usable for person identification. The original, obfuscated and reconstructed version of a sample frame can be seen in Figure \ref{fig:advobf}. In the obfuscated image, the persons have been transformed into silhouettes which are still correctly classified by the object detector but do not allow for identification. The results of the deobfuscator indicate that accurate reconstruction of the frames is impossible. However, further work is needed to prove the robustness of these claims.

\section{Experiments and results}
\label{sec:res}

All experiments have been implemented using the PyTorch~\cite{NEURIPS2019_9015} library. Unless stated otherwise we used the pretrained Faster R-CNN~\cite{ren2016faster} with ResNet50 backbone as our object detection model. We used street surveillance videos from the Avenues~\cite{abnormal2013lu} and ShanghaiTech Campus (STC)~\cite{liu2018ano_pred} datasets from which individual frames have been extracted and scaled to ($320 \times 200$).

\subsection{Detection accuracy}
As the datasets were originally intended for anomaly detection, we consider the detections of  a pretrained Faster R-CNN as ground truths. We report the Average Precision (AP) of person detections by Faster R-CNN on the obfuscated frames in Table \ref{table:acc}. We show only the person AP as our target task is people counting. To investigate generalization, we conducted experiments with obfuscators trained on the STC and the Avenues dataset. Further, to investigate the robustness across multiple cameras, we have split the STC dataset into two sets containing footage of different cameras (camera 1-8 and 9-13). The results indicate that models can generalize over different cameras, perspectives and illumination levels. However, the accuracy drops significantly when testing on a different dataset than it was trained on. This could be attributed to the large amounts of passengers in the Avenues dataset in comparison to the STC dataset. 

\begin{table}[h]
\centering
\caption{Person AP of Faster R-CNN with ResNet50 backbone on obfuscated versions of different datasets. }
\begin{tabular}{  l  c  c  }
\toprule
Train set & Test set  & Person AP (\%) \\ 
\midrule
 \multirow{3}{*}{STC 1-8 (train)} & STC 1-8 (test) & 87 \\ 
                                & STC 9-13 (all) & 86 \\ 
                                & Avenues (test) & 46 \\ \hline
\multirow{3}{*}{Avenues (train)} & STC 1-8 (test) & 35 \\ 
                        & STC 9-13 (all) & 40 \\ 
                        & Avenues (test) & 92 \\ 
\bottomrule
\end{tabular}
\label{table:acc}
\end{table}


A question one could pose is whether the obfuscator is overfitted to the object detection model it was trained with. Ideally, the obfuscator retains general features that all object detection models use. Table \ref{table:cross_model} shows the accuracy reached when the obfuscated frames are classified by another detector than the one used at training time. The accuracy remains fairly high when testing on different models. 
Even when testing on YOLO~\cite{yolov3}, a model with an entirely different architecture than the region based models, accuracy remains high. This generalized obfuscation means that users of third party object detection models can obfuscate their data without having to worry about the precise model used by the SaaS provider.

\begin{table}[h]
\centering
\caption{Relative cross model accuracy, ResNet50, MobileNetV3-Large and MobileNetV3-Large 320 indicate the backbone of a Faster R-CNN model. Obfuscators trained with a certain object detection model remain accurate in combination with different models. }
\resizebox{\columnwidth}{!}{\begin{tabular}{  l  c  c  c  c  c }
\toprule
\multirow{2}{*}{\hfil Trained on} & \multicolumn{5}{c|}{Tested on (Person AP \%)} \\ \cline{2-6}
   &  ResNet50 & MobileNetV3-Large & MobileNetV3-Large 320 &
    Mask R-CNN & YOLOv5s\\ \midrule
 ResNet50  & 92  & 77 & 28 & 83 & 89 \\ 
 MobileNetV3-Large  & 68  & 83 & 30 &65 & 81 \\ 
 MobileNetV3-Large 320  &  28 & 60 & 47  & 29 & 73 \\ 
 Mask R-CNN  & 81   & 64 & 22 & 82 & 86 \\ 
\bottomrule
\end{tabular}}
\label{table:cross_model}
\end{table}
\subsection{Obfuscation}
We also compare our approach with traditional techniques for obfuscation such as blurring, adding noise or quantizing the color information. We found that these techniques had a significantly worse accuracy-obfuscation trade-off. This is illustrated in Table \ref{table:obf_metrics} on frames from the Avenues dataset. To ensure a fair comparison, the parameter values of the obfuscation techniques were set to achieve similar average precision of person detection than our model. We quantitatively measure the quality of the obfuscation by calculating various similarity metrics between the original and obfuscated image:  the Structural Similarity Index Measure (SSIM)~\cite{ssim}, Mean Squared Error (MSE), Peak Signal to Noise Ratio (PSNR) and Normalized Mutual Information (NMI)~\cite{Studholme1999AnOI}. Except for the MSE, a lower score indicates a stronger obfuscation. 

\begin{table}[h]
\caption{Comparison of the similarity between original frames and frames obfuscated by various obfuscation techniques. Parameters of each technique were tuned to reach person AP similar to ours. Our technique results in better obfuscation.}

\centering
\begin{tabular}{  l   c  c  c  c  c }
\toprule
Metric & Ours (92.26\%) &  \thead{Blurring (91.77\%) \\ Kernel size: [3,3] } & \thead{Random Noise (87.73\%) \\ Noise factor: 2\%} &  \thead{Quantified (88.40\%) \\ 15 color values}  \\ 
\midrule
Visual & \includegraphics[width=0.15\linewidth]{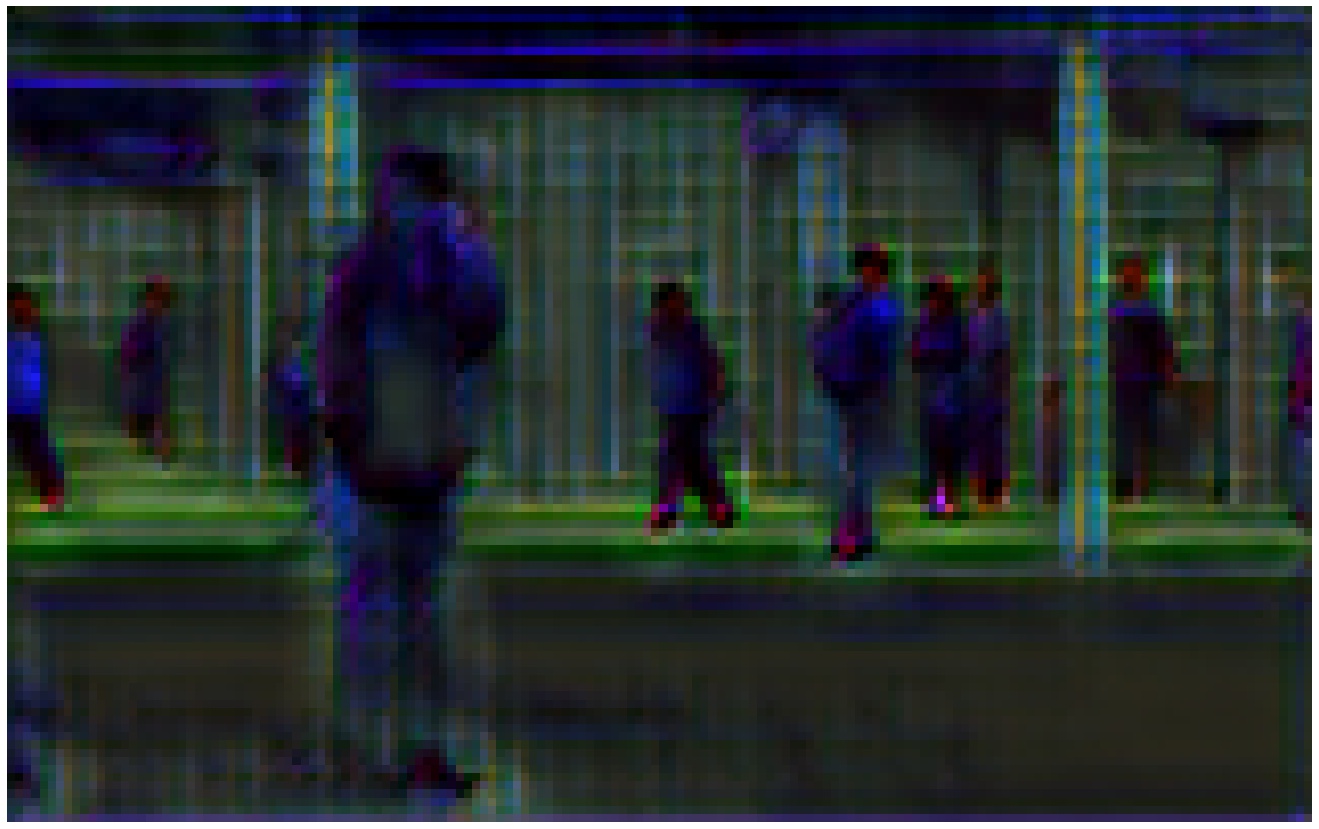} & \includegraphics[width=0.15\linewidth]{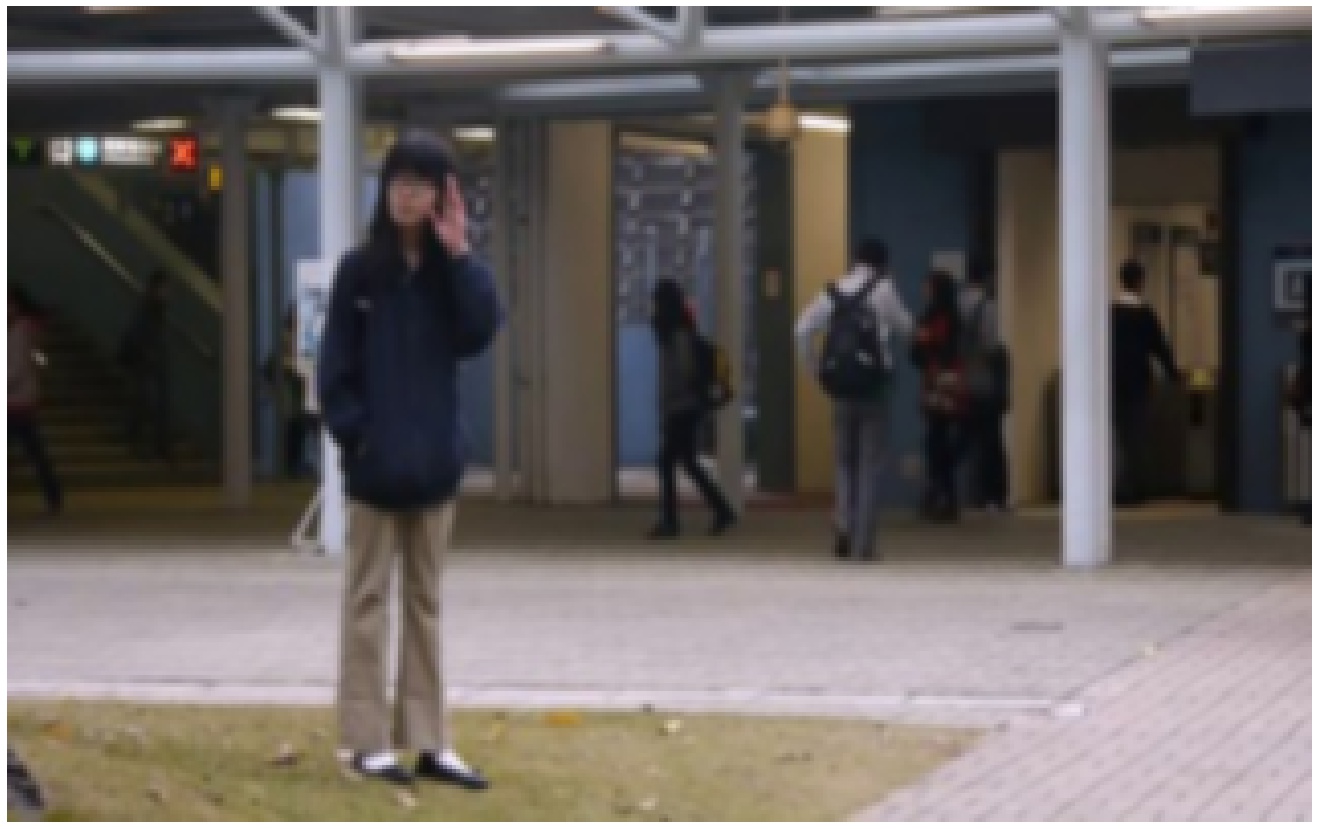} & \includegraphics[width=0.15\linewidth]{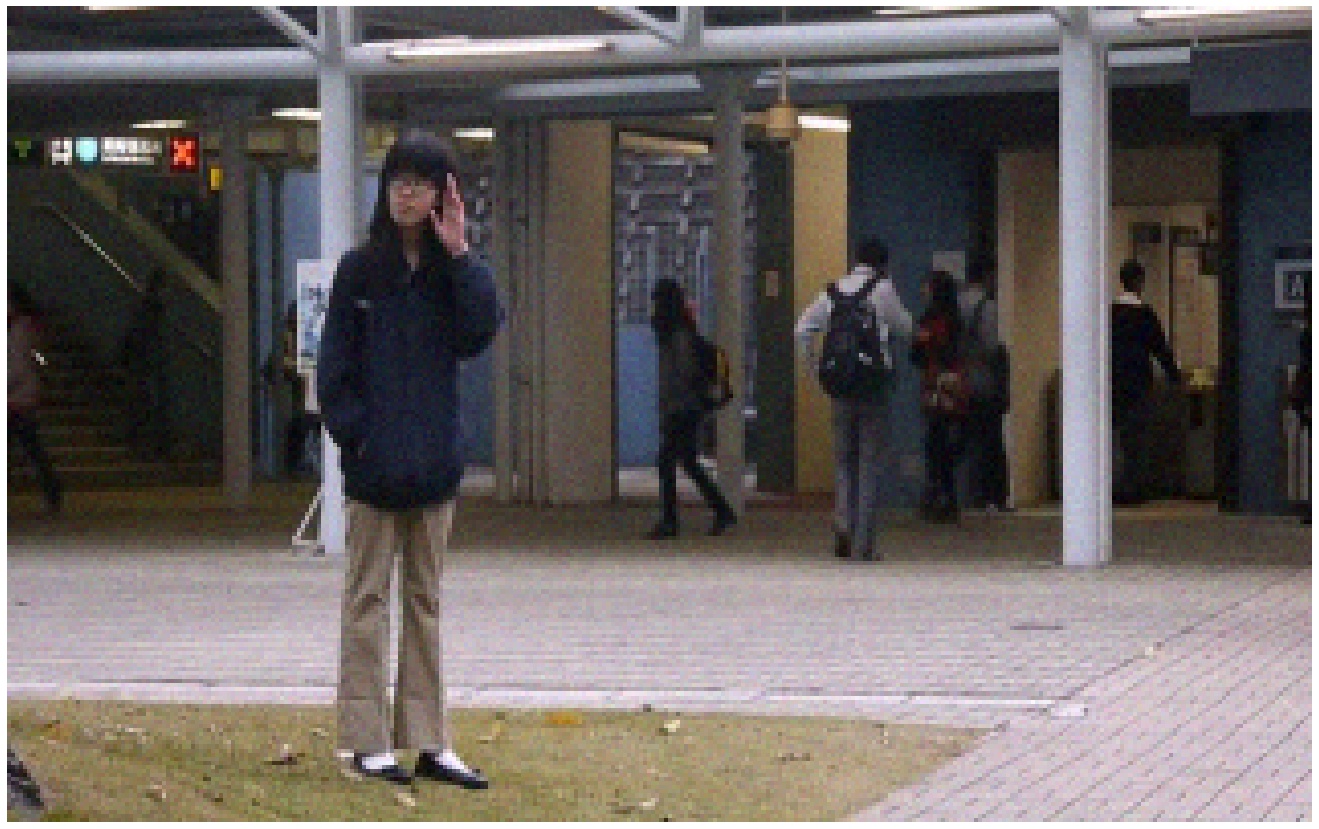}& \includegraphics[width=0.15\linewidth]{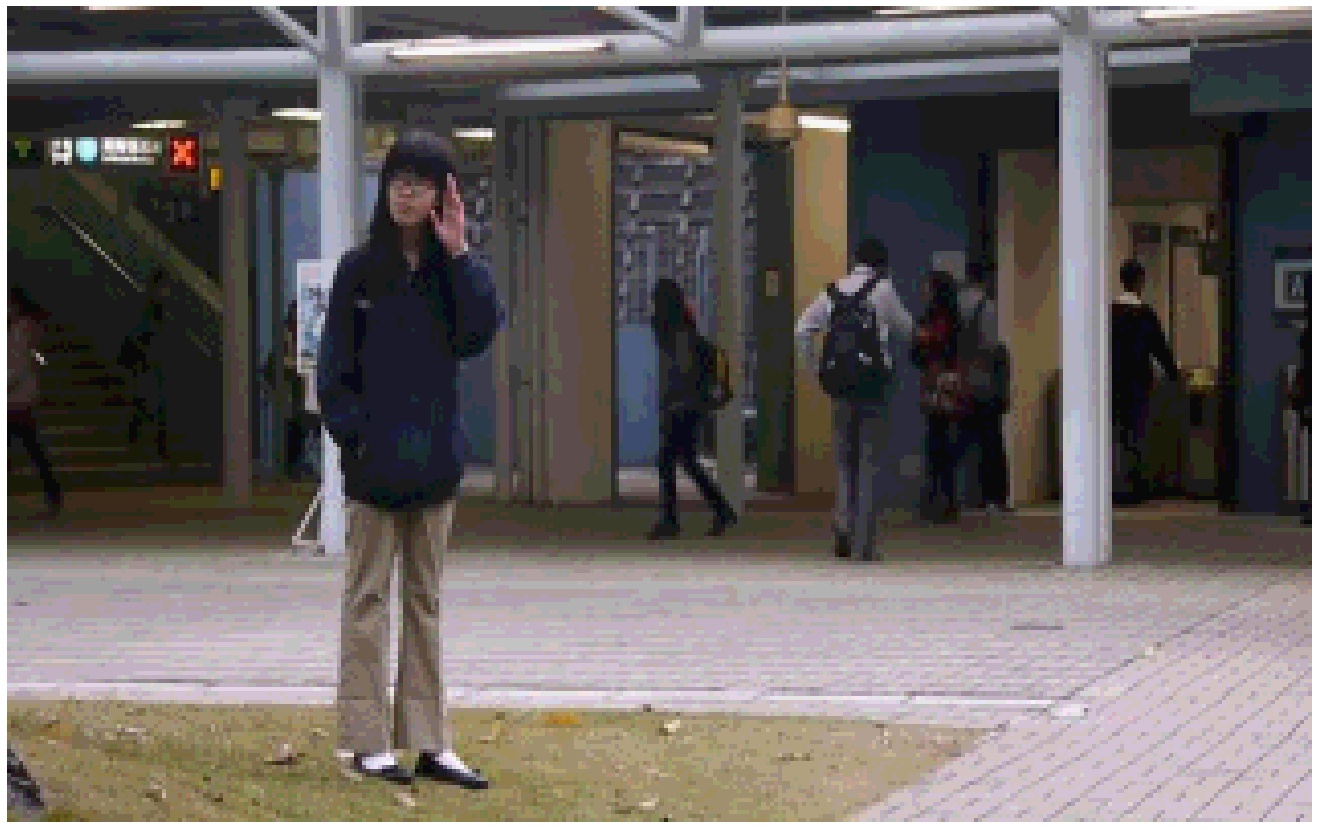} \\
 SSIM & 0.3998 & 0.9515 & 0.9512  & 0.9593  \\ 
MSE & 0.1191 & 0.0008 & 0.0004 & 0.0004  \\ 
 PSNR & 9.2494 & 30.9633 & 33.9794 & 34.2505   \\ 
NMI & 1.0247 & 1.3567 & 1.3531 & 1.5510  \\ 
\bottomrule
\end{tabular}
\label{table:obf_metrics}
\end{table}

\subsection{Computational efficiency}
Leaving aside the operational and economic benefits of using a object detector offered as a cloud service, from the perspective of energy efficiency  an obvious alternative is to run an object detector on the camera and only transfer the counts to the cloud. The footprint of a neural network is typically measured by means of the number of multiply-accumulate operations and 
the number of parameters to be stored in memory. 
In Table \ref{table:eff} we compare the size and computational cost of the obfuscator to those of two common object detection models. The obfuscator outperforms the evaluated object detection models in terms of multiply-accumulate operations (MACs) and parameters. By reducing the amount of convolutional filters in each layer, we can reduce the MACs of the obfuscator. The corresponding degradation in accuracy is shown in Figure~\ref{fig:macs}. Even after reducing the MACs with a factor 20, the detection accuracy is still above 70\%.
\begin{table}
	\begin{minipage}[t]{0.55\linewidth}
		\caption{Comparison between the obfuscator and object detection models (parentheses shows relative to ours): ours is considerably smaller.}
		\label{table:student}
		\centering
        \begin{tabular}{ l  c c  }
        \toprule
        Model & MACs (Gops) & Params (M)\\ 
        \midrule
         Obfuscator & 1.384 & 0.154  \\ 
         Faster R-CNN & 206.452 (149.2) & 41.75 (271.1) \\  
         YOLOv5s & 5.545 (4.0) & 7.277 (47.3) \\ \bottomrule
    \end{tabular}
    \label{table:eff}

	\end{minipage}\hfill
	\begin{minipage}[t]{0.45\linewidth}
	\vspace{0pt}

		\centering
        \includegraphics[width=\textwidth]{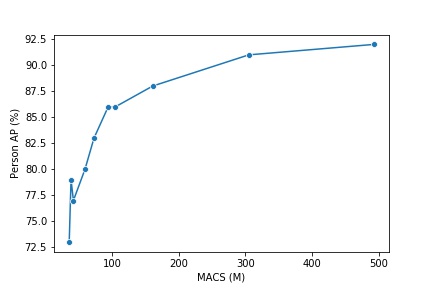}
		\captionof{figure}{Accuracy in function of MACs, the obfuscator can be significantly reduced without large accuracy loss.}
		\label{fig:macs}
	\end{minipage}
\end{table}

\section{Conclusion}
\label{sec:conclusion}
We have shown that our adversarial obfuscator structure can be successfully modified for the person detection problem. 
Our obfuscation of video frames results in a minimum degradation of person detection accuracy and generalizes well across camera viewpoints and object detector models. Furthermore, the obfuscation is suitable for deployment on the edge. 
Future work will focus on more formally proving the irreversibility of the obfuscation and on the robustness against function creep by an information-theoretic analysis of the mutual information between the obfuscated frame and the features required for other classification tasks.

\section*{Acknowledgements}
This research received funding from the Flemish Government under the ``Onderzoeksprogramma Artifici\"ele Intelligentie (AI) Vlaanderen'' programme.

{
\small

\bibliography{main}
}

\appendix

\section{Appendix}

\subsection{Training details}
We can describe the loss function of the deobfuscator, where $X$ is the original image; $\hat{X}$ the output of the deobfuscator and MSE the mean squared error as follows:
$$\mathcal{L}_{deobf}(X, \hat{X}) = MSE(X, \hat{X}) $$
The loss of the obfuscator can be described with the object detection model's loss function $\mathcal{L}_{obj}$ and ground truth labels $Y$ as:
$$\mathcal{L}_{obf}(X, Y, \hat{X}) = \mathcal{L}_{obj}(X, Y) - \mathcal{L}_{deobf}(X, \hat{X})$$

We alternatingly trained the obfuscator and deobfuscator for 30 epochs total starting with a learning rate of $ 1 \times 10^{-2}$ and $ 1 \times 10^{-3}$ respectively. Every ten epochs we divided the obfuscator's learning rate by 100 and the deobfuscator's by ten. The AdamW~\cite{loshchilov2019decoupled} optimizer was used for both with a weight decay of $ 1 \times 10^{-4}$. 

\subsection{Obfuscation examples}

On the next pages we show multiple frames of the Avenues and ShanghaiTech Campus datasets with their obfuscations and deobfuscations.
\begin{figure}[h]
  \begin{subfigure}{\linewidth}
  \includegraphics[width=\linewidth]{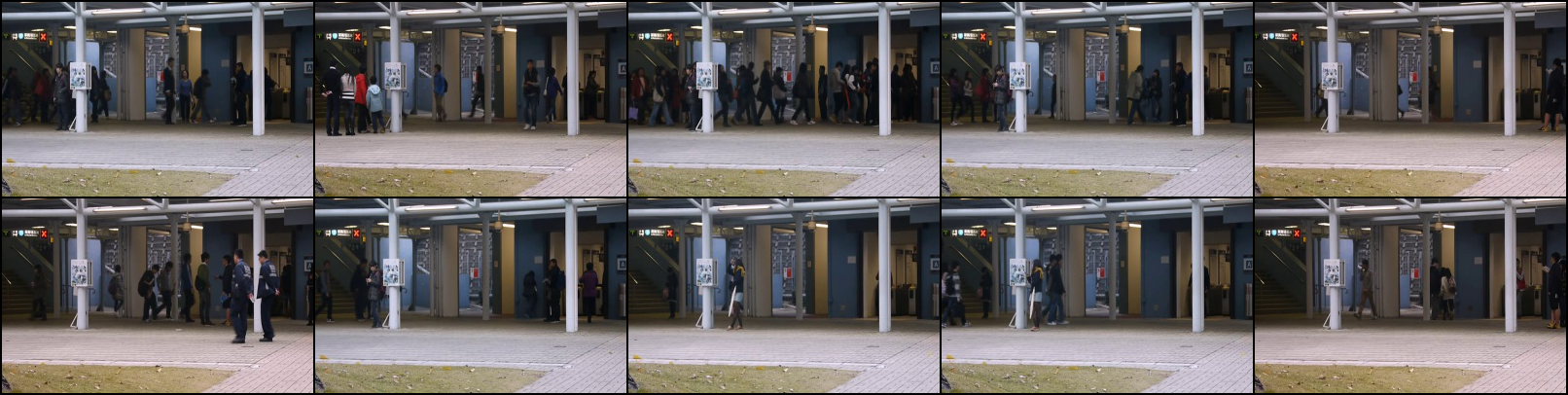}%
  \label{fig:obf_a}
  \caption{Original images}
  \end{subfigure} \par \medskip

  \begin{subfigure}{\linewidth}
  \includegraphics[width=\linewidth]{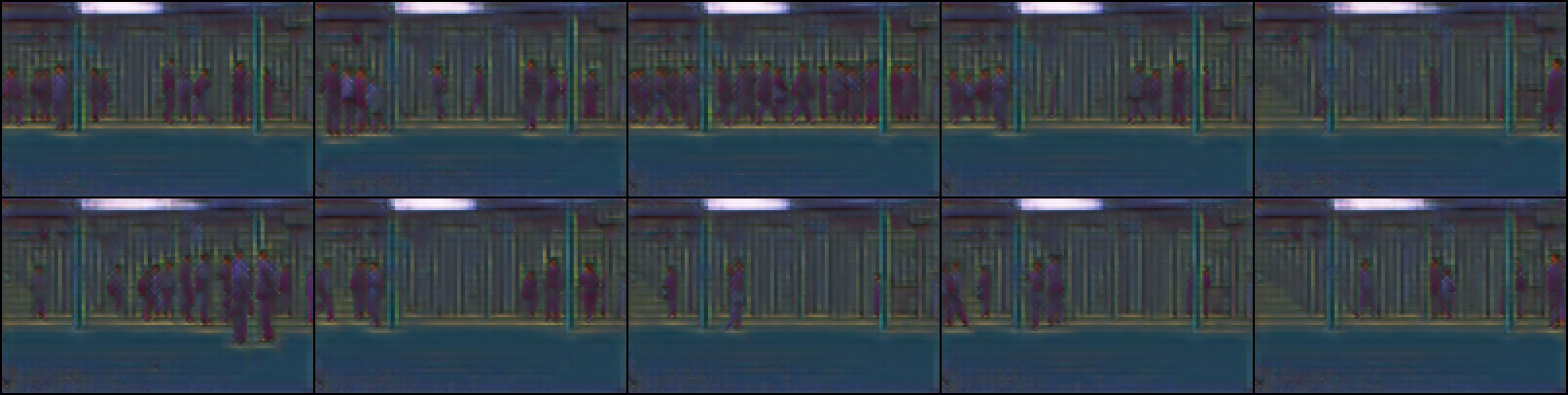}%
  \label{fig:obf_a}
  \caption{Obfuscated images}
  \end{subfigure} \par \medskip

  \begin{subfigure}{\linewidth}
  \includegraphics[width=\linewidth]{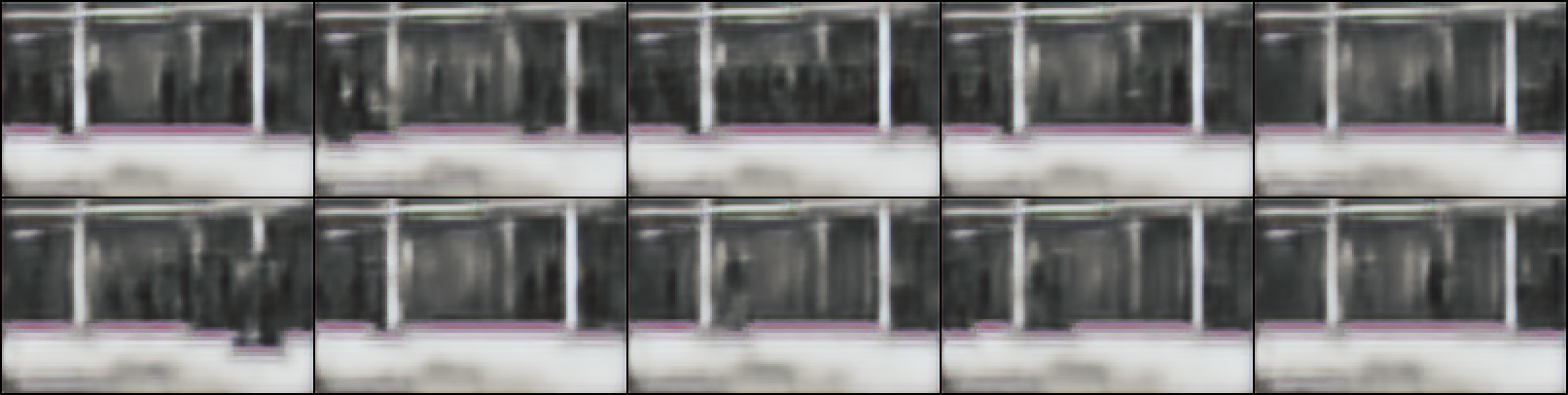}%
  \label{fig:obf_a}
  \caption{Reconstructed images}
  \end{subfigure} \par \medskip
  \caption{Obfuscation results on the Avenues dataset}
  \label{fig:obf_av}
\end{figure}

\begin{figure}[h]
  \begin{subfigure}{\linewidth}
  \includegraphics[width=\linewidth]{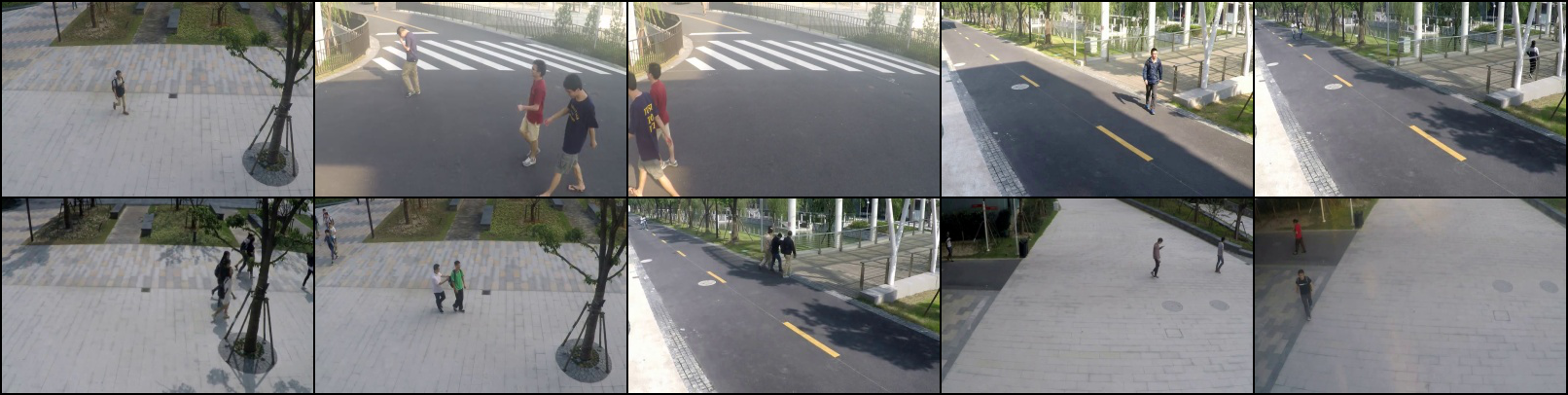}%
  \label{fig:obf_a}
  \caption{Original images}
  \end{subfigure} \par \medskip

  \begin{subfigure}{\linewidth}
  \includegraphics[width=\linewidth]{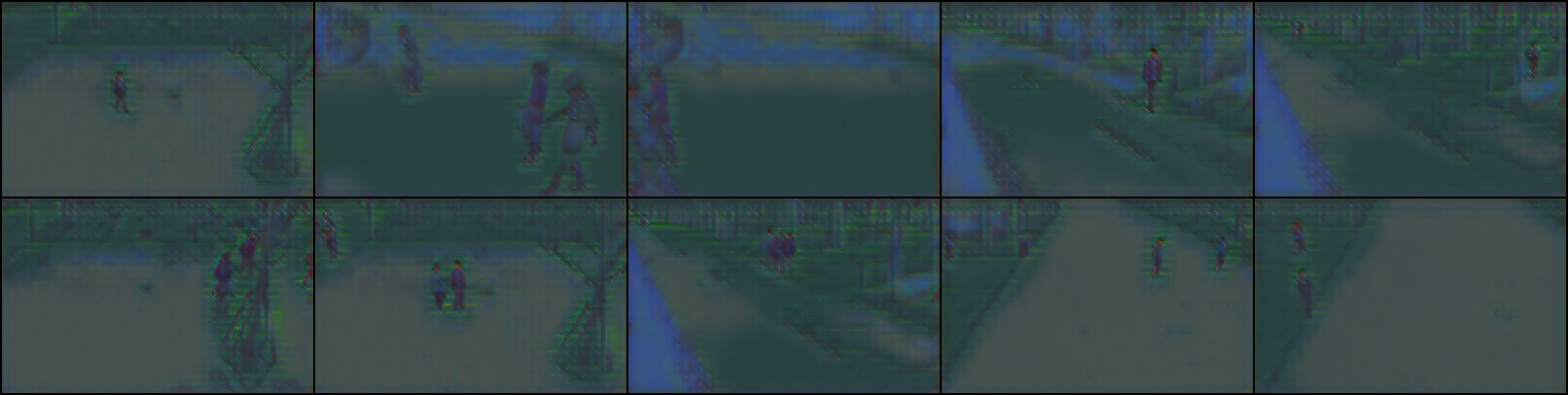}%
  \label{fig:obf_a}
  \caption{Obfuscated images}
  \end{subfigure} \par \medskip

  \begin{subfigure}{\linewidth}
  \includegraphics[width=\linewidth]{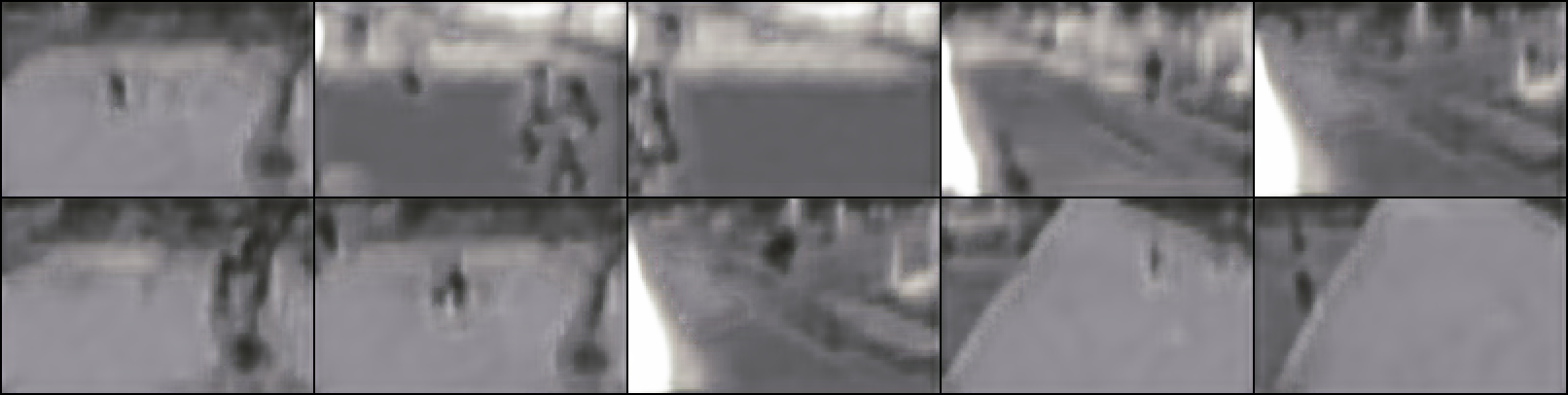}%
  \label{fig:obf_a}
  \caption{Reconstructed images}
  \end{subfigure} \par \medskip
  \caption{Obfuscation results on the ShanghaiTech Campus dataset}
  \label{fig:obf_stc}
\end{figure}
\end{document}